\documentclass{article}

\PassOptionsToPackage{numbers, sort&compress}{natbib}
\usepackage[preprint]{neurips_2023}

\usepackage[utf8]{inputenc} 
\usepackage[T1]{fontenc}    
\usepackage{url}            
\usepackage{booktabs}       
\usepackage{amsfonts}       
\usepackage{nicefrac}       
\usepackage{microtype}      
\usepackage{enumitem}
\usepackage{wrapfig} 
\usepackage{graphicx}
\usepackage{subfigure}
\usepackage{amsmath}
\usepackage{amssymb}
\usepackage{booktabs}
\usepackage{bbm}
\usepackage{enumitem}
\usepackage[dvipsnames,svgnames,x11names,table]{xcolor} 
\usepackage{pifont}
\usepackage{bbding}

\usepackage{wrapfig}
\usepackage{multirow}
\usepackage{transparent}
\usepackage{tikz}
\usepackage[table]{xcolor}
\definecolor{pretty-blue}{RGB}{0, 113, 188}
\definecolor{brown}{RGB}{201, 104, 71}

\definecolor{softred}{RGB}{231,76,60}     
\definecolor{softgreen}{RGB}{46,204,113}  
\definecolor{softyellow}{RGB}{241,196,15} 
\definecolor{bggreen}{RGB}{105,169,78}     
\usepackage{makecell}

\usepackage[colorlinks=True,
            linkcolor=blue,
            anchorcolor=blue,  
            pagebackref,
            urlcolor=blue,
            citecolor=teal,
            ]{hyperref}

\title{\ \ \ General OCR Theory:  Towards OCR-2.0 \\via a Unified End-to-end Model}

\author{
Haoran Wei$^{1,}$\thanks{Equal contribution} \ ,
Chenglong Liu$^{3,*}$, Jinyue Chen$^{3}$, Jia Wang$^{1}$, Lingyu Kong$^{3}$,  \bf{Yanming Xu}$^{1}$, \\ \bf{Zheng Ge}$^{1}$, \bf{Liang Zhao}$^{1}$, \bf{Jianjian Sun}$^{1}$, \bf{Yuang Peng}$^{4}$, \bf{Chunrui Han}$^{2}$,  \bf{Xiangyu Zhang}$^{1,2}$\\
\\
$^{1}$StepFun  \ \ \ $^{2}$Megvii Technology \\ $^{3}$University of Chinese Academy of Sciences    \ \ $^{4}$Tsinghua University \\
\url{https://github.com/Ucas-HaoranWei/GOT-OCR2.0} \\
}

\begin{document}

\maketitle

\begin{tikzpicture}[remember picture,overlay,shift={(current page.north west)}]
\node[anchor=north west, xshift=3.25cm, yshift=-2.75cm]{\scalebox{1}[1]{\includegraphics[width=2.5cm]{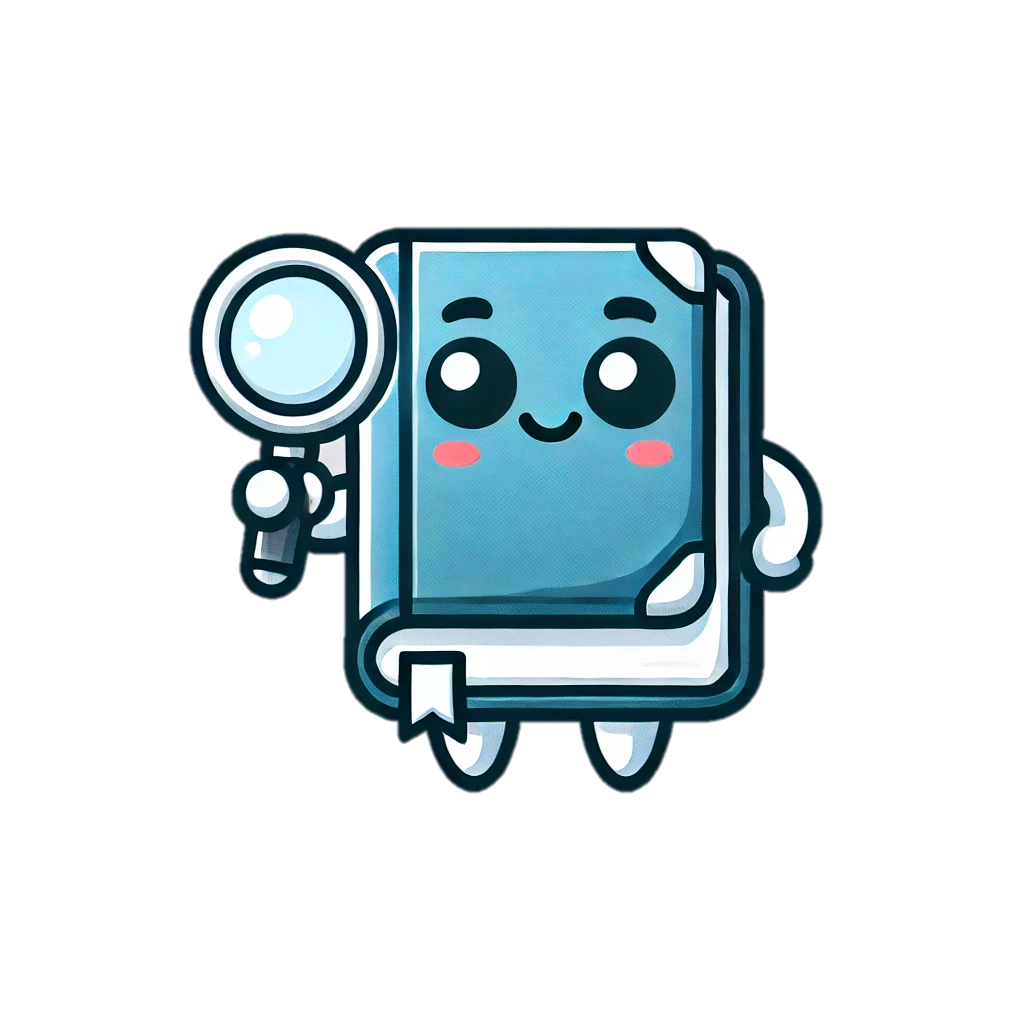}}};
\end{tikzpicture}




\begin{abstract}
Traditional OCR systems (OCR-1.0) are increasingly unable to meet people's usage due to the growing demand for intelligent processing of man-made optical characters. In this paper, we collectively refer to all artificial optical signals (e.g., plain texts, math/molecular formulas, tables, charts, sheet music, and even geometric shapes) as "characters" and propose the \textbf{G}eneral \textbf{O}CR \textbf{T}heory along with an excellent model, namely GOT, to promote the arrival of OCR-2.0. The GOT, with 580M parameters, is a unified, elegant, and end-to-end model, consisting of a high-compression encoder and a long-contexts decoder.  As an OCR-2.0 model, GOT can handle all the above "characters" under various OCR tasks. On the input side, the model supports commonly used scene- and document-style images in slice and whole-page styles. On the output side, GOT can generate plain or formatted results (markdown/tikz/smiles/kern) via an easy prompt. Besides, the model enjoys interactive OCR features, i.e., region-level recognition guided by coordinates or colors. Furthermore, we also adapt dynamic resolution and multi-page OCR technologies to GOT for better practicality.  In experiments, we provide sufficient results to prove the superiority of our model.

\end{abstract}


\section{Introduction}
\label{intro}
Optical Character Recognition (OCR) is a widely used technology that extracts the characters embedded in an optical image into an editable format.  Typical OCR systems~\cite{paddleocrv2_du2021pp} in the OCR-1.0 era are mainly designed based on a multi-modular pipeline style, commonly including element detection, region cropping, and character recognition parts. Each module is prone to falling into local optima, making the whole system incur high maintenance costs. Moreover, traditional OCR methods have insufficient general ability, reflected as different OCR-1.0 networks usually designed for different sub-tasks. 
Nevertheless, choosing a suitable one from diverse OCR models for a special task is always inconvenient for users.

\begin{figure}[!h]
\centering
\includegraphics[width=1.0\textwidth]{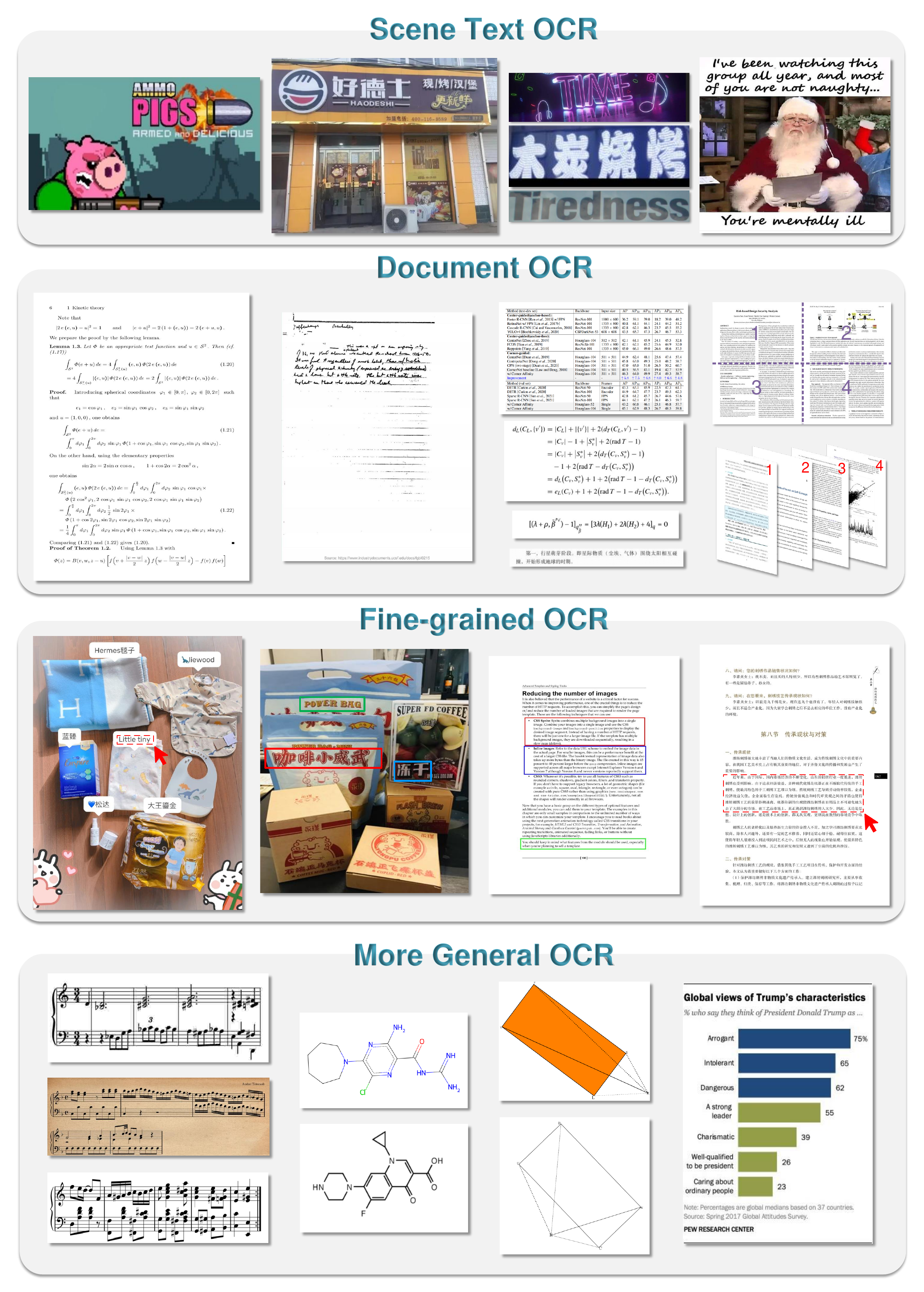}
\caption{On the input side, GOT supports various optical image types, such as commonly used photographs and documents. Besides, as a general OCR-2.0 model, GOT can handle more tasks, e.g., sheet music, molecular formulas, easy geometric shapes, charts, etc. Moreover, the model can adapt to region-focus OCR, high-resolution OCR, and multiple-page OCR. GOT mainly supports English and Chinese and can control the structure results (Mathpix markdown/tikz/smiles/kern) via a prompt.}
\label{fig:intro}
\end{figure}

In the past year, Large Vision Language models (LVLMs)~\cite{GPT4,llava,Qwen-VL,wei2023vary,ye2023mplug,chen2024far_intervl1.5,liu2024textmonkey} have developed rapidly and showcased impressive performance. As a highly anticipated ability, the OCR performance of current LVLMs is continuously improving. Based on CLIP~\cite{radford2021learning}, LLaVA~\cite{llava} naturally
acquires the English OCR ability after the instruct tuning phase. To lift the OCR accuracy and support other languages, e.g., Chinese, Qwen-VL~\cite{Qwen-VL} unfreezes its image encoder (a CLIP-G) and uses lots of OCR data in its stage-two training. Innovatively, Vary~\cite{wei2023vary} generates a new high-resolution OCR vision vocabulary paralleling the CLIP branch to deal with document-level dense OCR. By contrast, InternVL-1.5~\cite{chen2024far_intervl1.5} and other models~\cite{liu2024textmonkey,ye2023ureader} utilize a sliding window manner to crop the whole image into multiple sub-patches for high-resolution OCR. Hence, a consensus is that optical character perception and recognition are the foundation of text-driven image understanding, drawing many researchers to pay more attention to LVLMs' OCR booster.

However, the popular designs of LVLMs may not be suitable for diverse OCR tasks for the following reasons: 1) The conflicts between perception and reasoning.  LVLMs mainly focus on visual reasoning performance, e.g., VQA~\cite{TextVQA,DocVQA}, because that is what the LLM excels at.  To quickly obtain the QA-gain benefits from LLMs, most LVLMs~\cite{llava,ye2023mplug,BLIP2} align image tokens to text ones. However, it is unreasonable to do this for pure perception OCR tasks, especially high-density text scenes, because each aligned vision token (biased towards text token) cannot compress enough characters. Imagine how wasteful it is to use thousands of image tokens, e.g., the image-cropping manner~\cite{chen2024far_intervl1.5,liu2024llavanext}, to encode an equal amount of optical characters (e.g., texts within only an A4-PDF page). 2) High iteration and deployment costs. LVLM often enjoys billions of parameters, leading to the post-training and deployment costs being too high. Generally speaking, for LVLMs, fine-tuning is not enough once we want to add a new OCR pattern, e.g., a new language, instead of enough GPU resources for pre-training. However, rerunning the pre-training with billions of parameters, only to introduce a new OCR feature, is also wasteful.

Accordingly, we propose the general OCR theory, i.e., OCR-2.0, to break the bottlenecks of both traditional and LVLM manners on OCR tasks. We think that a model of OCR 2.0 should have the following essential characteristics:
\begin{itemize}[leftmargin=*]
\item \textbf{End-to-end.} Compared to OCR-1.0 models with complex procedures, the OCR-2.0 model should enjoy a unified and end-to-end architecture to ensure lower maintenance costs. It is cool that a beginner can quickly master the entire OCR system in the 2.0 era.
\item \textbf{Low training and inference costs.} The OCR-2.0 model should not be a chatbot, like LVLM, that focuses on reasoning tasks. Its focus should be on strong perception and recognition of optical characters, so it needs a reasonable number of model parameters in exchange for lower training and inference costs. 
\item \textbf{Versatility.} The OCR-2.0 model's other important point is versatility, including recognizing more general artificial optical ``characters'', e.g., sheet music, charts, geometric shapes, etc. Besides, the model 
should support the output format with stronger readability, e.g., \LaTeX{}/Markdown format for formulas and tables.

\end{itemize}

Based on the proposed general OCR theory, we present a primary OCR-2.0 model (GOT) to bridge the gap between OCR-1.0 models and people's higher optical character processing demands. In architecture, we adopt the unsophisticated encoder-decoder paradigm for the model.
Specifically, GOT enjoys a high compression rate encoder to transfer the optical image to tokens as well as a long context length decoder to output the corresponding OCR results. The encoder has approximately 80M parameters posing 1024$\times$1024 input size which is enough to deal with commonly used photo/document input styles. Each input image will be compressed to tokens with 256$\times$1024 dimensions. The decoder of GOT, with 0.5B parameters, supports 8K max length tokens to ensure it can tackle long-context scenarios. We devise an effective and efficient training strategy for GOT, which can be divided into three procedures, i.e., decoupled pre-training of the encoder, joint-training of the encoder with a new decoder, and further post-training of the decoder. Besides, to further lift the practicality of GOT, we additionally adapt the fine-grained OCR feature for better interactivity, dynamic resolution strategy for ultra-high-resolution images (e.g., over 2K), and the multi-page OCR technology to alleviate the problem of difficulty in breaking pages in PDF image-text pairs (e.g., page breaks in \textit{.tex} files).  To support each training stage, we do many data engines for synthetic data production, which is the key to the success of GOT and will be described in detail in this paper. The main input data format supported by our model can be seen in Figure~\ref{fig:intro}.

As a model for envisioning OCR-2.0, GOT demonstrates promising performance in our experiments in various OCR tasks. We hope the proposed simple and elegant GOT can draw more researchers to invest in the research of OCR-2.0. Of course, the path to OCR-2.0 is still long and GOT also enjoys much improvement room, such as supporting more languages, more general artificial signals, and more complex geometries. In this new era led by LVLMs, we are convinced that the pure OCR model is not over, it may even be a new beginning.

\section{Related Work}

\subsection{Traditional OCR}
Optical Character Recognition (OCR) is a classic research topic that aims to convert the image's optical contents into an editable format for further downstream processing. Traditional OCR systems, called OCR-1.0, typically use a framework that is assembled from multiple expert modules. For instance, to handle diverse optical characters, the OCR system~\cite{paddleocrv2_du2021pp} is usually developed by integrating several domain expert networks, such as layout analysis~\cite{zhong2019publaynet}, text detection~\cite{tian2016detecting,liao2017textboxes,liao2022real_dbnet,zhou2017east,lyu2018multi,liu2019curved,wang2020contournet,zhang2021adaptive}, region extraction, and contents recognition~\cite{lecun1998gradient_mnist,graves2006connectionist,li2023trocr}. The reason for using such a pipeline scheme is that the text recognition module (the OCR part) failed to scale up successfully, which can only deal with the image format of small slices, resulting in the entire OCR process being in the form of first detecting texts/cropping regions, and then recognizing the results within the slice. However, a system with complicated procedures may suffer potential systematic errors and high maintenance costs. Although some OCR-1.0 models, e.g., Nougat~\cite{blecher2023nougat} can directly process documents at the whole page level, they are often designed and trained for a specific sub-task, leading to unsatisfactory general ability. In the OCR-1.0 era, one inconvenient thing is that we usually need to switch different models according to various OCR needs.


\subsection{LVLM-driven OCR}
Large Vision-Language Models (LVLMs)~\cite{llava,Qwen-VL,wei2023vary,ye2023mplug,chen2024far_intervl1.5,liu2024textmonkey,liu2024focus_fox} have attracted lots of attention in the AI-community due to their powerful generalization capabilities. For the current LVLMs owning perception-reasoning comprehensive capacity, the OCR ability has become a hot spot with the increasing demand for text-driven visual understanding. Most LVLMs' OCR capabilities come from the ready-made CLIP~\cite{radford2021learning}, especially those that freeze CLIP encoder~\cite{llava} to complete the entire LVLM training. For such models, the vanilla CLIP, mainly with English scene text knowledge, is the bottleneck for the OCR performance to out-of-domain tasks, such as other languages or documents. Some other LVLMs~\cite{ye2023mplug,Qwen-VL} choose to unfreeze the encoder and freeze the LLM for training to enhance the CLIP-encoder and align the image tokens to text ones. These models will face the problem of low optical character compression rate, as it is difficult for frozen LLM to decode too much text from an aligned image token. To alleviate this problem, some models~\cite{chen2024far_intervl1.5,liu2024textmonkey,ye2023ureader} adopt a sliding window manner to decompose input images into smaller patches. Although this dynamic resolution approach is highly effective in processing high-resolution input images, e.g., PDF, it will result in excessive image tokens and limit the max length of the generated OCR result to some extent.

\section{General OCR Theory}
In this work, we propose the general OCR theory, i.e., OCR-2.0 (as expounded in Section~\ref{fig:intro}) to promote the development of the OCR field. Based on the proposed new theory, we present a novel OCR model (GOT). In this section, we will introduce the technical details of our model, including the framework, multi-stage training strategy, and the corresponding data engines. 

\begin{figure}[t]
\centering
\includegraphics[width=0.9\textwidth]{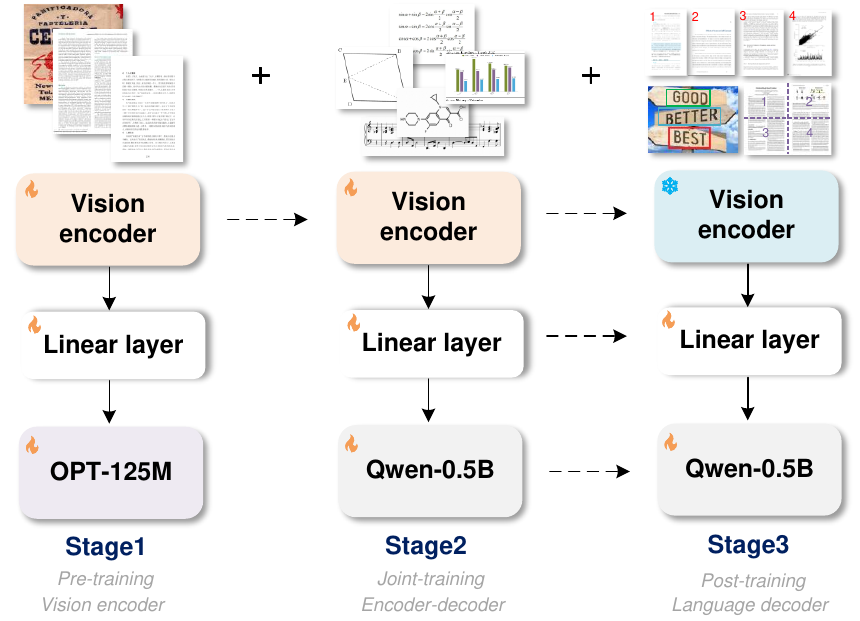}
\caption{The framework of the proposed GOT. Stage 1: We pre-train the vision encoder using a tiny OPT-125M to adapt the OCR tasks efficiently. Stage 2: GOT is built by connecting the vision encoder to Qwen-0.5B and sufficient OCR-2.0 knowledge of more general optical characters is used in this stage. Stage 3: No modification of the vision encoder is required, and GOT is customized to new character recognition features.}
\label{fig:framework}
\end{figure}

\subsection{Framework}
As illustrated in Figure~\ref{fig:framework}, GOT comprises three modules, i.e., an image encoder, a linear layer, and an output decoder. The linear layer acts as the connector to map the channel dimension between the vision encoder and the language decoder. We utilize three main steps in optimizing the whole GOT model. First, we conduct the pure text recognition task to pre-train the vision encoder. To lift training efficiency and save GPU resources, we choose a tiny decoder to pass gradients to the encoder. In this stage, we feed images containing scene texts and manual images containing document-level characters into the model to allow the encoder to gather the two most commonly used characters' encoding abilities.
In the next stage, we form the architecture of GOT by connecting the trained vision encoder to a new larger decoder. We prepare lots of more general OCR data (\textit{e.g.}, sheet music, math/molecular formulas, and geometric shapes) to scale up the OCR-2.0 knowledge for this stage. 
In the final stage, we intend to improve the generalization and applicability of GOT further. Specifically, fine-grained and muti-crop/page synthetic data are generated and added for GOT to support region prompt OCR~\cite{liu2024focus_fox}, huge image OCR, and batched PDF OCR features.

\subsection{Pre-train the OCR-earmarked Vision Encoder}
As aforementioned, GOT enjoys the encoder-decoder structure. Inspired by the LVLMs design, the decoder can be initialized by a well-trained language model. However, we did not find a suitable pre-trained encoder for an OCR-2.0 model, so we must train one ourselves.  
We hope the new OCR encoder can work well on commonly used scene and document text recognition in various input shapes (both slices and whole pages).
\subsubsection{The Vision Encoder Generation.}
The encoder structure we selected is VitDet ~\cite{li2022exploring_vitdet} (base version with about 80M parameters) due to its local attention can greatly reduce the computational cost of high-resolution images. We follow the Vary-tiny setting~\cite{wei2023vary} to design the last two layers of the encoder, which will transfer a 1024$\times$1024$\times$3 input image to 256$\times$1024 image tokens. 
Then, these image tokens are projected into language model (OPT-125M~\cite{OPT}) dimension via a 1024$\times$768 linear layer. Unlike the Vary encoder which only focuses on a single document task under a relatively unitary input shape, we incorporated natural scenes and cropped slices during our pre-training. In the pre-processing stage, images of each shape are directly resized to 1024$\times$1024 squares, as square shapes can be used to adapt to images of various aspect ratios with a compromise.



\subsubsection{Data Engine Towards Encoder Pre-training} \label{pretrain}
In such an encoder pre-training stage, we use about 5M image-text pairs, including 3M scene text OCR data and 2M document OCR data. Their acquisition methods are as follows:

For the natural scene data, the English and Chinese images are sampled from Laion-2B~\cite{schuhmann2022laion5b} and Wukong~\cite{gu2022wukong} datasets, respectively. Then, the pseudo ground truth in these diverse real scenes is captured using PaddleOCR~\cite{paddleocrv2_du2021pp} tools. Overall, we obtain 2M dat with half in Chinese and half in English. For text ground truth, we perform two types of processing: 1) remove the bounding box and combine each text content in order from top to bottom and left to right. 2) crop the text region from the original image according to the bounding box and save it as image slices. The later method 2) allows us to obtain another 1M slice-type image-text pairs. 

For the document-level data, we first collect open-source PDF-style files from the Common Crawl and employ the Fitz Python package to extract corresponding dense text content. In such a process, we gain 1.2M full-page PDF-style image-text pairs and 0.8M image slice data. The slice data, including line- and paragraph-level, is cropped from the PDF image via the parsed bounding box.

\subsection{Scaling Up the OCR-2.0 Knowledge via  Multi-task Joint-training}


\subsubsection{The Final Architecture of GOT}
After the pre-training step of the vision encoder, we connect it to a larger language model with more powerful capabilities to build the final architecture of GOT. Here, we adopt the Qwen~\cite{qwen} with 500M parameters as the decoder because it has a relatively small number of parameters while incorporating prior knowledge of multiple languages. The dimension of the connector (i.e., the linear embedding layer) is adjusted into 1024$\times$1024 to align with the input channels of the Qwen-0.5B. Hence, GOT enjoys the seamless encoder-decoder paradigm with about 580M parameters in total, which is more computationally resource-friendly and easier to deploy on a consumer-grade GPU with 4G memory. The high compression rate (1024$\times$1024 optical pixels to 256 image tokens) of the encoder saves a lot of token space for the decoder to generate new tokens. Meanwhile, the satisfactory decoding context length (we use about 8K max-length) of the decoder ensures that the GOT can effectively output OCR results under dense scenes.

\subsubsection{Data Engine for Joint-training}
\begin{figure}[!t]
\centering
\includegraphics[width=1.0\textwidth]{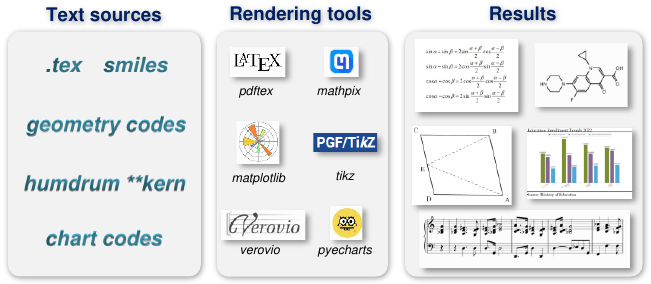}
\caption{We use six rendering tools to run data engines to make the GOT work well on diverse OCR tasks. We utilize the \LaTeX{} for tables, Mathpix-markdown-it for math/molecular formulas, Tikz for simple geometric shapes, Verovio for sheet music, and Matplotlib/Pyecharts for charts, respectively.  }
\label{fig:render}
\end{figure}

To inject sufficient OCR-2.0 knowledge into GOT, instead of the above-mentioned plain OCR data, we carefully explore several synthesis methods and data engines in this stage, as shown in Figure~\ref{fig:render}. We will delve into the details of each type of synthetic data in the following paragraphs.

\paragraph{Plain OCR data.}
We use 80\% of the data mentioned in Section~\ref{pretrain} as plain OCR data. To further enhance the robustness of GOT, we also add the handwritten text recognition sub-task, which involves various styles of handwriting from letters and diaries in different languages. 
We collect the Chinese CASIA-HWDB2~\cite{Teklia_CASIA-HWDB2-line}, English IAM~\cite{Teklia_IAM-line}, and Norwegian NorHand-v3~\cite{Teklia_NorHand-v3-line} datasets to meet our requirements. For the original image-text pairs with the line-level slice format, 6$\sim$8 pairs are grouped and randomly pasted into a blank document page to achieve longer-text handwriting recognition and improve training efficiency.

\paragraph{Mathpix-markdown formatted data.}
Preserving the optical content format is critical to maintaining strong readability for the output results, especially for mathematical formulas and tables. To this end, we use multiple approaches to gather as much formatted data as possible. The details of data collection and production are as follows:

\begin{itemize}[leftmargin=*]
\item \textbf{Math formulas.} We crawl a large number of \LaTeX{} source \textit{.tex} files on Arxiv and extract about 1M formula fragments from them. Next, we transfer the formula sources to Mathpix format and use the Chorme-driver to call Mathpix-markdown-it tool to render the sources to HTML format. We then convert the HTML files to SVGs and save them as PNG images. We find that this rendering method is more than 20$\times$ faster than directly using the \LaTeX{}.

\item \textbf{Molecular formulas.} We first download the \textit{ChEMBL$\_$25} file that contains 2M smile sources. Then we use the Mathpix-markdown-it tool and \textit{rdkit.Chem} package to gather about 1M of molecular formula image-text pairs.

\item \textbf{Table}. From the crawled \textit{.tex} files, we extract about 0.3M table sources and render them into images. Instead of Mathpix-markdown-it, we directly utilize the  \LaTeX{} as the rendering tool due to its better rendering effects for advanced tables. 

\item \textbf{Full page data.} Using the Nougat~\cite{blecher2023nougat} method, we obtain about 0.5M English markdown PDF-text pairs. Besides, following Vary~\cite{wei2023vary,wei2024small_varytoy}, we gather another 0.5M Chinese markdown pairs.  We transfer their contents to Mathpix format.  Furthermore, we additionally add 0.2M in-house data, which is directly labeled using Mathpix, including books, papers, and financial reports.

\end{itemize}

\paragraph{More general OCR data.} We hope GOT can deal with more general optical artificial ``characters''. Accordingly, we collect three related challenging tasks and generate the corresponding data. They are sheet music, geometric shapes, and charts, respectively.

\begin{itemize}[leftmargin=*]
\item \textbf{Sheet music.} Music is a precious part of the cultural heritage and optical music recognition plays an important role in achieving automatic recognition and transcription of sheet music~\cite{calvo2020understanding_OMR,rios2024sheet_music}. We choose the GrandStaff~\cite{rios2023end_GrandStaff} dataset as the source to render. The dataset of polyphonic music scores provides the \textit{Humdrum **kern} transcriptions from the excerpts of music. In addition to the existing approximately 10w image-text samples, we also extract some text samples to re-render via the Verovio Python Package.  We mainly add new backgrounds from white to real paper styles and randomly add the title and author information. Note that we only render single-system sheet music due to we don't have professionals in the relevant field and we do not know how to assemble single-system sheets to a full page. After rendering, we collect about 0.5M samples.


\item \textbf{Geometric shape.} Geometry is a key capability of LVLMs and is a necessary step towards AGI.
GOT is expected to transform optical geometric elements into TikZ~\cite{mertz2007graphics_tikz} text format. TikZ contains some concise commands to produce basic geometric elements and they can be compiled using \LaTeX{}.
We employ TikZ-style points and lines and use the simplest point-line spatial relationship to construct simple basic geometric shapes (\textit{e.g.}, circles, rectangles, triangles, and combined shapes) as well as simple function curves (\textit{e.g.}, straight lines, parabolas, ellipses, hyperbolas, and so on). Through this method, we obtained approximately 1M geometric Tikz data. Of course, the geometric rendering is complicated, and our current work is only a preliminary attempt. GOT can only recognize basic geometry at present, yet we believe that with the development of synthetic data technology and OCR-2.0, future models will be able to identify complex geometric shapes.

\item \textbf{Chart.} Charts are crucial in data visualization and data analysis of several research fields. The proposed GOT refers to the chart structural extraction sub-task as ``Chart OCR'', which converts the visual knowledge (\textit{e.g.}, title, source, x-title, y-title, and values) on the chart image into an editable output with a table/Python-dict format. Following OneChart~\cite{chen2024onechart}, the chart image-text pairs are rendered using Matplotlib and Pyecharts tools. Because GOT is only an OCR model, we don't need the elements of the chart synthesized to be semantically related. Thus, we just randomly extract entity texts (for the title, source, x-title, y-title, etc) from the open-access NLP corpus. The numerical values are random numbers under a controlled distribution.  Through this method, we obtained 2M chart data, with half from Matplotlib and half from Pyecharts.

\end{itemize}

\subsection{Customizing New OCR Features by Post-training the Decoder}
After compressing the general visual information of the diverse OCR-2.0 optical signals via the above two steps, GOT is ready to perform image-level OCR tasks in various scenarios. Based on this perceptually savvy vision encoder, GOT can be easily tuned to meet the users' needs for input and output. Here, we customize GOT to enable three new features, i.e., fine-grained, multi-page, and dynamic resolution OCR, by only post-training the decoder part. 


\subsubsection{Fine-grained Data Engine for Interactive OCR.}
As a high-interactivity feature, fine-grained OCR~\cite{liu2024focus_fox} is the region-level visual perception controlled by spatial coordinates or colors. The user can add box coordinates (box-guided OCR) or color text (color-guided OCR) in the question prompt to request recognition within the region of interest (RoI), avoiding the output of other irrelevant characters. 
For the natural fine-grained OCR, the source images and annotations are from opensource datasets, including RCTW~\cite{shi2017icdar2017_RCTW}, ReCTS~\cite{liu2019icdar_ReCTS}, and ShopSign~\cite{zhang2019shopsign}, and COCO-Text~\cite{veit2016coco_text} dataset. The datasets mentioned above provide the text bounding boxes, so we can use them to produce fine-grained (region/color prompt) OCR data directly.
For the document-level fine-grained OCR, following Fox~\cite{liu2024focus_fox}, we filter out those with the scanned format in the downloaded PDF files and parse the left part using Python packages (Fitz/PDFminer). We record the page-level images, bounding boxes of each line/paragraph, and the corresponding texts to produce the ground truth of the box-guided OCR sub-task. For such a task, each coordinate value is first normalized and then magnified 1000 times. For the color-guided task, we choose the most commonly used colors (red, green, and blue) as the frame colors and draw them via the corresponding bounding box on the original image. Overall, we gather about 60w samples.

\subsubsection{Multi-crop Data Engine for Ultra-large-image OCR.}
GOT supports 1024$\times$1024 input resolution, which is enough for commonly used OCR tasks, e.g., scene OCR or A4-page PDF OCR. However, dynamic resolution is required for some scenes with huge images, such as two-page PDF horizontal stitching (commonly occurring when reading papers). Thanks to our high compression rate encoder, the dynamic resolution of GOT is achieved under a large sliding window (1024$\times$1024), ensuring that our model can complete extreme resolution OCR tasks with acceptable image tokens. We use the InternVL-1.5~\cite{chen2024far_intervl1.5} cropping method with tiles max to 12. The ultra-resolution images are synthesized using the single-page PDF data mentioned above, including horizontal and vertical stitching. Through this method, we obtained a total of 50w image-texts pairs.

\subsubsection{Multi-page Data Engine for Batched PDF-file OCR.}
For OCR tasks, it is reasonable to use a ``for loop'' for multi-page processing. We introduce the multi-page OCR (without ``for loop'') feature for GOT due to some formatted PDF data making it difficult to break pages (to obtain text that is completely incompatible with each page) to further scale up, such as \textit{.tex} in Arxiv. We hope that with GOT, researchers no longer have to worry about PDF ground truth page breaks (e.g., Nougat~\cite{blecher2023nougat}), as they can train on multiple pages directly. To realize such a feature, we randomly sample 2-8 pages from our Mathpix formatted PDF data and join them together to form a single round OCR task. Each selected page contains text that is less than 650 tokens, to ensure that the overall length does not exceed 8K. In total, we generate about 20w multi-page OCR data, most of which are interlaced between Chinese and English pages.

\section{Experiments}

\subsection{Implement Details}
We use 8$\times$8 L40s GPUs to train GOT.  In the pre-training stage,  we optimize all model parameters with a global batch size of 128 and train for 3 epochs. We utilize the AdamW~\cite{AdamW} optimizer and a cosine annealing scheduler~\cite{loshchilov2016sgdr} with a start learning rate of 1e-4. The max token length in this stage is set to 4096. In the joint-training stage, we put the max token length to 6000 and train the model with the same optimizer settings as stage 1 for 1 epoch. In the last post-training stage, we expand the max token length to 8192 to allow the model to support multi-patch/page OCR features. In this stage, the beginning learning rate is 2e-5, and the epoch is set to 1. 

During each train-data process, 80\% of the data from the previous stage is sampled for the following stage to ensure that the basic ability does not degrade when adding new features. 

\begin{table*}[ht]
\scriptsize
\centering
\setlength{\tabcolsep}{4pt}
{
\begin{tabular}{lccccccccccccc}
\toprule[.9pt]
\multirow{2}{*}{{{Method}}} & \multirow{2}{*}{Size} & \multicolumn{2}{c}{Edit Distance$\downarrow$} & \multicolumn{2}{c}{F1-score$\uparrow$}  &  \multicolumn{2}{c}{Precision$\uparrow$} & \multicolumn{2}{c}{Recall$\uparrow$} & \multicolumn{2}{c}{BLEU$\uparrow$} & \multicolumn{2}{c}{METEOR$\uparrow$}  \\  
\cmidrule(rl){3-4} \cmidrule(rl){5-6}  \cmidrule(rl){7-8}  \cmidrule(rl){9-10}  \cmidrule(rl){11-12} \cmidrule(rl){13-14} & & en & zh & en & zh & en & zh & en & zh & en & zh & en & zh \\ 
\midrule
UReader~\cite{ye2023ureader} & 7B & 0.718 & - & 0.344 & - & 0.296 & -& 0.469 & - & 0.103 & - & 0.287 & - \\
LLaVA-NeXT~\cite{liu2024llavanext} & 34B & 0.430 &-& 0.647 &- & 0.573 & -& 0.881 & -&  0.478 &- & 0.582 & -\\
{InternVL-ChatV1.5\cite{chen2024far_intervl1.5}} & 26B & 0.393 & 0.265 & 0.751 & 0.816 & 0.698 & 0.784 & 0.917 & 0.866 & 0.568 & 0.622 & 0.663 & 0.717 \\ 
{Nougat~\cite{blecher2023nougat}} & 250M & 0.255 & -& 0.745 & -& 0.720 & -& 0.809 & -& 0.665 & -& 0.761 & -\\ 
TextMonkey~\cite{liu2024textmonkey} & 7B & 0.265 & - & 0.821 & - & 0.778 & -& 0.906 & - & 0.671 & -& 0.762 & -\\
DocOwl1.5~\cite{hu2024mplugdocowl1.5} & 7B & 0.258 & - & 0.862 & - & 0.835 & -& 0.962 & - & 0.788 & - & 0.858 & - \\
Vary~\cite{wei2023vary} & 7B & 0.092 & 0.113 & 0.918 & 0.952 & 0.906 & 0.961 & 0.956 & 0.944 & 0.885 & 0.754 & 0.926 & 0.873 \\
{Vary-toy~\cite{wei2024small_varytoy}} & 1.8B & 0.082 & 0.142 & 0.924 & 0.914 & 0.919 & 0.928 & 0.938 & 0.907 & 0.889 & 0.718 & 0.929 & 0.832 \\  
Qwen-VL-Plus~\cite{Qwen-VL} & - & 0.096 & 0.121 & 0.931 & 0.895 & 0.921 & 0.903 & 0.950 & 0.890 & 0.893 & 0.684 & 0.936 & 0.828 \\
{Qwen-VL-Max~\cite{Qwen-VL}} & >72B & 0.057 & 0.091 & {0.964} & 0.931 & 0.955 & 0.917 & \cellcolor{blue!3}\bf{0.977} & 0.946 & {0.942} & 0.756 & \cellcolor{blue!3}\bf{0.971} & 0.885 \\ 
Fox~\cite{liu2024focus_fox} & {1.8B} & {0.046} & {0.061} & 0.952  & {0.954}  & {0.957} & {0.964} & 0.948 & {0.946} & 0.930 & {0.842} &  0.954 & {0.908} \\ 

\textbf{GOT} & 580M & \cellcolor{blue!3}\textbf{0.035} & \cellcolor{blue!3}\textbf{0.038} & \cellcolor{blue!3}\textbf{0.972} & \cellcolor{blue!3}\textbf{0.980} & \cellcolor{blue!3}\textbf{0.971} & \cellcolor{blue!3}\textbf{0.982} & \cellcolor{white}0.973 & \cellcolor{blue!3}\textbf{0.978} & \cellcolor{blue!3}\textbf{0.947} & \cellcolor{blue!3}\textbf{0.878} & \cellcolor{white}0.958 & \cellcolor{blue!3}\textbf{0.939} \\
\bottomrule[.9pt]
\end{tabular}
}
\caption{Performance comparison of dense English (en) and Chinese (zh) OCR on document-level pages. The results of other models are from the previous work~\cite{liu2024focus_fox}.} 
\label{tab:en_page_ocr}
\end{table*}

\subsection{Main Results}
In this section, we verify the performance of GOT on 5 different OCR tasks, including 1) plain document OCR; 2) scene text OCR;  3) fine-grained document OCR; 4) formatted (Mathpix markdown) document OCR; 5) more general character OCR. Note that the test data for each benchmark undergoes strict text similarity filtering to ensure that it is not included in the training data. Sources of each test benchmark and model performance analysis are as follows.

\subsubsection{Plain document OCR performance}
We use the open-source Fox~\cite{liu2024focus_fox} benchmark to test the performance of GOT on both Chinese and English PDF OCR. The metrics we used are those commonly in OCR tasks, i.e., edict distance, F1-score, precision, recall, BLEU, and METEOR.  Due to the lengthy text of the document, we use word-level segmentation to calculate each indicator. As shown in Table~\ref{tab:en_page_ocr}, with only 580M, GOT achieves advanced performance on pure text OCR in the document, proving the excellent PDF text perception and recognition ability.

\begin{table*}[ht]
\scriptsize
\centering
\setlength{\tabcolsep}{4.2pt}
{
\begin{tabular}{lccccccccccccc}
\toprule[.9pt]
\multirow{2}{*}{{{Method}}} & \multirow{2}{*}{Size} & \multicolumn{2}{c}{Edit Distance$\downarrow$} & \multicolumn{2}{c}{F1-score$\uparrow$}  &  \multicolumn{2}{c}{Precision$\uparrow$} & \multicolumn{2}{c}{Recall$\uparrow$} & \multicolumn{2}{c}{BLEU$\uparrow$} & \multicolumn{2}{c}{METEOR$\uparrow$}  \\  
\cmidrule(rl){3-4} \cmidrule(rl){5-6}  \cmidrule(rl){7-8}  \cmidrule(rl){9-10}  \cmidrule(rl){11-12} \cmidrule(rl){13-14} & & en & zh & en & zh & en & zh & en & zh & en & zh & en & zh \\ 
\midrule
UReader~\cite{ye2023ureader} & 7B & 0.568 & - & 0.661 & - & 0.843 & -& 0.569 & - & 0.258 & - & 0.488 & - \\
LLaVA-NeXT~\cite{liu2024llavanext} & 34B & 0.499 &-& 0.558 &- & 0.637 & -& 0.538 & -&  0.379 &- & 0.678 & -\\
TextMonkey~\cite{liu2024textmonkey} & 7B & 0.331 & - & 0.743 & - & 0.827 & -& 0.710 & - & 0.521 & -& 0.728 & -\\
DocOwl1.5~\cite{hu2024mplugdocowl1.5} & 7B & 0.334 & - & 0.788 & - & 0.887 & -& 0.751 & - & 0.525 & - & 0.708 & - \\
{InternVL-ChatV1.5\cite{chen2024far_intervl1.5}} & 26B & 0.267 & 0.123 & 0.834 & 0.913 & \cellcolor{blue!3}\textbf{0.942} & \cellcolor{blue!3}\textbf{0.934} & 0.790 & 0.902 & 0.587 & 0.588 & 0.744 & 0.876 \\ 
{Qwen-VL-Max~\cite{Qwen-VL}} & >72B & 0.182 & 0.168 & 0.881 & 0.867 & 0.891 & 0.878 & 0.888 & 0.873 & 0.586 & 0.572 & 0.848 &  0.845\\ 
\textbf{GOT} & 580M & \cellcolor{blue!3}\textbf{0.112} & \cellcolor{blue!3}\textbf{0.096} & \cellcolor{blue!3}\textbf{0.926} & \cellcolor{blue!3}\textbf{0.928} & 0.934 & {0.914} & \cellcolor{blue!3}\textbf{0.927} & \cellcolor{blue!3}\textbf{0.954} & \cellcolor{blue!3}\textbf{0.676} & \cellcolor{blue!3}\textbf{0.641} & \cellcolor{blue!3}\textbf{0.896} & \cellcolor{blue!3}\textbf{0.928} \\
\bottomrule[.9pt]
\end{tabular}
}
\caption{Performance of English (en) and Chinese (zh) OCR for scene texts.}
\label{tab:en_scene_ocr}
\end{table*}

\subsubsection{Scene text OCR performance}

We collect 400 natural images, half in Chinese and half in English, as the scene text OCR benchmark. All the ground truth in this benchmark are manually corrected. Because the text in the scene image is relatively short, we use character-level segmentation to calculate various metrics. As shown in Table~\ref{tab:en_scene_ocr}, we can see that GOT also works well on natural images, demonstrating the model's excellent performance on most basic OCR tasks (both document and scene texts).

\subsubsection{Formatted document OCR performance}

Converting the optical PDF image to a markdown-like format is an important feature of an OCR model. To verify this ability of GOT, we carefully prepare 90 pages of samples as a high-quality benchmark. The benchmark, containing both Chinese and English document pages, is first generating pseudo-labels via Mathpix, and then manually correcting for errors. In Table~\ref{tab:mathpix}, we can see the single-scale (1024$\times$1024) GOT can yield satisfactory results. When we use multi-crop inference, the performance of GOT is further lifted especially on formulas and tables with small texts. The results prove the effectiveness of GOT on documents with formatted outputs. Besides, the dynamic resolution scheme is a good choice when processing higher-resolution images.

\begin{table*}[!t]
\scriptsize
\centering
\setlength{\tabcolsep}{5pt}
{
\begin{tabular}{l|ccccccccc}
\toprule[.9pt]
& Types & Edit Distance$\downarrow$ &F1-score$\uparrow$  & Precision$\uparrow$ & {Recall$\uparrow$} & {BLEU$\uparrow$} &{METEOR$\uparrow$}  \\  
\midrule

\multirow{8}{*}{\thead[l]{\makecell{Markdown\\document}}} 

& \cellcolor{gray!10}\textbf{single:} & \cellcolor{gray!10}& \cellcolor{gray!10}& \cellcolor{gray!10}& \cellcolor{gray!10}&\cellcolor{gray!10} &\cellcolor{gray!10} \\
&All text & 0.097 & 0.942 & 0.944 & 0.942 & 0.877 & 0.876 \\
&Formula & 0.269 & 0.749 & 0.771 & 0.751 & 0.512 & 0.716 \\
&Table & 0.254 & 0.867 & 0.857 & 0.897 & 0.756 & 0.760 \\
&\cellcolor{gray!10}\textbf{muti-crop:} & \cellcolor{gray!10}& \cellcolor{gray!10}& \cellcolor{gray!10}& \cellcolor{gray!10}& \cellcolor{gray!10}&\cellcolor{gray!10}  \\
&All text &0.086 & 0.953 & 0.948 & 0.960 & 0.896 & 0.903   \\
&Formula & 0.159& 0.865 & 0.858 & 0.882 & 0.628 & 0.828   \\
&Table & 0.220& 0.878 & 0.861 & 0.919 & 0.779 & 0.811   \\

\midrule
\midrule

\multirow{2}{*}{\thead[l]{\makecell{Geneal}}} 
&Sheet music &  0.046 & 0.939 & 0.963  & 0.939  & 0.900  & 0.923 \\
&Geometry & 0.061 & 0.884 & 0.882  & 0.888 & 0.766 & 0.882  \\

\bottomrule[.9pt]
\end{tabular}
}
\caption{Performances of formatted document (Chinese/English) and more general OCR. Single means the input is the vanilla image and multi-crop represents the dynamic resolution strategy.}
\label{tab:mathpix}
\vspace{-1.em}
\end{table*}

\subsubsection{Fine-grained OCR performance}
We report the fine-grained OCR metrics of GOT. As shown in Table~\ref{tab:boxline}, the GOT is overall better than Fox~\cite{liu2024focus_fox} on both the bounding box-based and color-based referential OCR tasks, indicating that our model enjoys excellent interactive OCR capabilities.  

\begin{table*}[ht]
\scriptsize
\centering
\setlength{\tabcolsep}{5pt}
{
\begin{tabular}{lcccccccccc}
\toprule[.9pt]
\multirow{3}{*}{\textbf{Metrics}} & \multicolumn{5}{c}{\textbf{English}} & \multicolumn{4}{c}{\textbf{Chinese}}  \\ 
\cmidrule(rl){2-6} \cmidrule(rl){7-10} & \multicolumn{3}{c}{{region}} & \multicolumn{2}{c}{{color}}  & \multicolumn{2}{c}{{region}} & \multicolumn{2}{c}{{color}}\\ 
\cmidrule(rl){2-4} \cmidrule(rl){5-6} \cmidrule(rl){7-8} \cmidrule(rl){9-10}  & DocOwl1.5~\cite{hu2024mplugdocowl1.5} & Fox~\cite{liu2024focus_fox} &  GOT & Fox~\cite{liu2024focus_fox} & GOT & Fox~\cite{liu2024focus_fox} & GOT & Fox~\cite{liu2024focus_fox} & GOT \\
\midrule
Edit Distance $\downarrow$  & 0.435 & {0.059} & \cellcolor{blue!3}\textbf{0.041} & 0.064  & \cellcolor{blue!3}\textbf{0.034} & 0.042 &  \cellcolor{blue!3}\textbf{0.033} & 0.114 & \cellcolor{blue!3}\textbf{0.040} \\ 
F1-score $\uparrow$ & 0.670 & {0.957} & \cellcolor{blue!3}\textbf{0.970} & 0.940  & \cellcolor{blue!3}\textbf{0.966} & 0.955 & \cellcolor{blue!3}\textbf{0.965} &  0.884 & \cellcolor{blue!3}\textbf{0.957} \\
Precision $\uparrow$ & 0.886 & {0.962} & \cellcolor{blue!3}\textbf{0.973} & 0.942 & \cellcolor{blue!3}\textbf{0.970}  & 0.966 & \cellcolor{blue!3}\textbf{0.974} &  0.902 & \cellcolor{blue!3}\textbf{0.969} \\ 
Recall $\uparrow$  & 0.617 & {0.955} & \cellcolor{blue!3}\textbf{0.969} & 0.942 & \cellcolor{blue!3}\textbf{0.964} & 0.947 & \cellcolor{blue!3}\textbf{0.958} & 0.873  & \cellcolor{blue!3}\textbf{0.948} \\ 
BLEU $\uparrow$ & 0.478 & {0.914} & \cellcolor{blue!3}\textbf{0.926} & 0.868  & \cellcolor{blue!3}\textbf{0.910} & 0.885 & \cellcolor{blue!3}\textbf{0.898} & 0.778  & \cellcolor{blue!3}\textbf{0.884} \\
METEOR $\uparrow$ & 0.569 & {0.955} & \cellcolor{blue!3}\textbf{0.966} & 0.938  & \cellcolor{blue!3}\textbf{0.961} & 0.934 & \cellcolor{blue!3}\textbf{0.942} & 0.848  & \cellcolor{blue!3}\textbf{0.931} \\
\bottomrule[.9pt]
\end{tabular}

}

\caption{Comparison of fine-grained document OCR.}
\label{tab:boxline}
\vspace{-1.em}
\end{table*}

\begin{table*}[!h]
\label{baseline}
\centering
\scriptsize
\setlength{\tabcolsep}{5pt}
{
\begin{tabular}{llccccccccc}
\toprule[.9pt]
\textbf{\textbf{~}} & Metric  & \makecell{Deplot\\(1.3B)~\cite{liu-2022-deplot}} & \makecell{UniChart\\(0.26B)~\cite{masry2023unichart}} & \makecell{ChartVLM\\(7.3B)~\cite{xia2024chartx}} &\makecell{GPT-4V\\(>100B)~\cite{GPT4}} & \makecell{Qwen-VL\\(>72B)~\cite{Qwen-VL}}&\makecell{\textbf{GOT}\\(0.58B)}  \\ 
\midrule
\multirow{3}{*}{\thead[l]{\makecell{ChartQA-SE}}}
& AP@strict  &0.614  & 0.423 & 0.718 & 0.504 &  0.586   &  \cellcolor{blue!3}\textbf{0.747}  \\ 
 & AP@slight  & 0.709 & 53.18 & 0.814 & 0.606 & 0.685   &  \cellcolor{blue!3}\textbf{0.845} \\ 
 & AP@high & 0.729 &  0.560  & 0.842 & 0.643 &  0.727   &  \cellcolor{blue!3}\textbf{0.867} \\ 
\midrule
\multirow{3}{*}{\thead[l]{\makecell{PlotQA-SE}}}
& AP@strict   &0.031 & 0.105 & 0.038 & 0.073 &  0.005  & \cellcolor{blue!3}\textbf{0.133} \\ 
 & AP@slight  &16.49 & 0.260 & 0.468 &  0.194 & 0.042   & \cellcolor{blue!3}\textbf{0.596}\\ 
 & AP@high &26.50 &  0.269  & 0.540 &  0.223 &  0.120   & \cellcolor{blue!3}\textbf{0.640}\\ 
\bottomrule[.9pt]
\end{tabular}

}
\caption{Performance comparisons on number-centric chart OCR.}
\label{tab:chart}
\end{table*}

\vspace{-.2em}

\subsubsection{More general OCR performance}
We utilize the sheet music, geometry, and chart benchmarks to verify GOT's more general OCR performance.
For the first two tasks, we separately render 100 and 180 additional samples as benchmarks, and as can be seen in Table~\ref{tab:mathpix}, GOT still performs well on these new OCR tasks. For chart OCR, we use structure-extraction version~\cite{chen2024onechart} ChartQA~\cite{masry2022chartqa} and PlotQA~\cite{methani2020plotqa} as benchmarks. In Table~\ref{tab:chart}, the chart OCR ability of GOT is even much better than the chart-specific models and popular LVLMs. All results demonstrate the effectiveness of our model on more general OCR tasks.

\section{Conclusion}
This paper presents a primary OCR-2.0 model that is structurally simpler than OCR-1.0 systems, focuses more on pure OCR tasks than LVLMs, and enjoys superior performance. OCR-2.0 integrates various pan-OCR tasks into one model and is a valuable research direction in model design, data engineering, and application scenarios. We want the simple, elegant, effective, and promising GOT OCR-2.0 model to attract more attention to such a task.

\section{Appendix}
In this section, we provide sufficient output results of GOT to show its outstanding OCR performance. We also demonstrate the format of the corresponding input prompt for different types of OCR tasks. 

\begin{figure}[!h]
\centering
\includegraphics[width=1.0\textwidth]{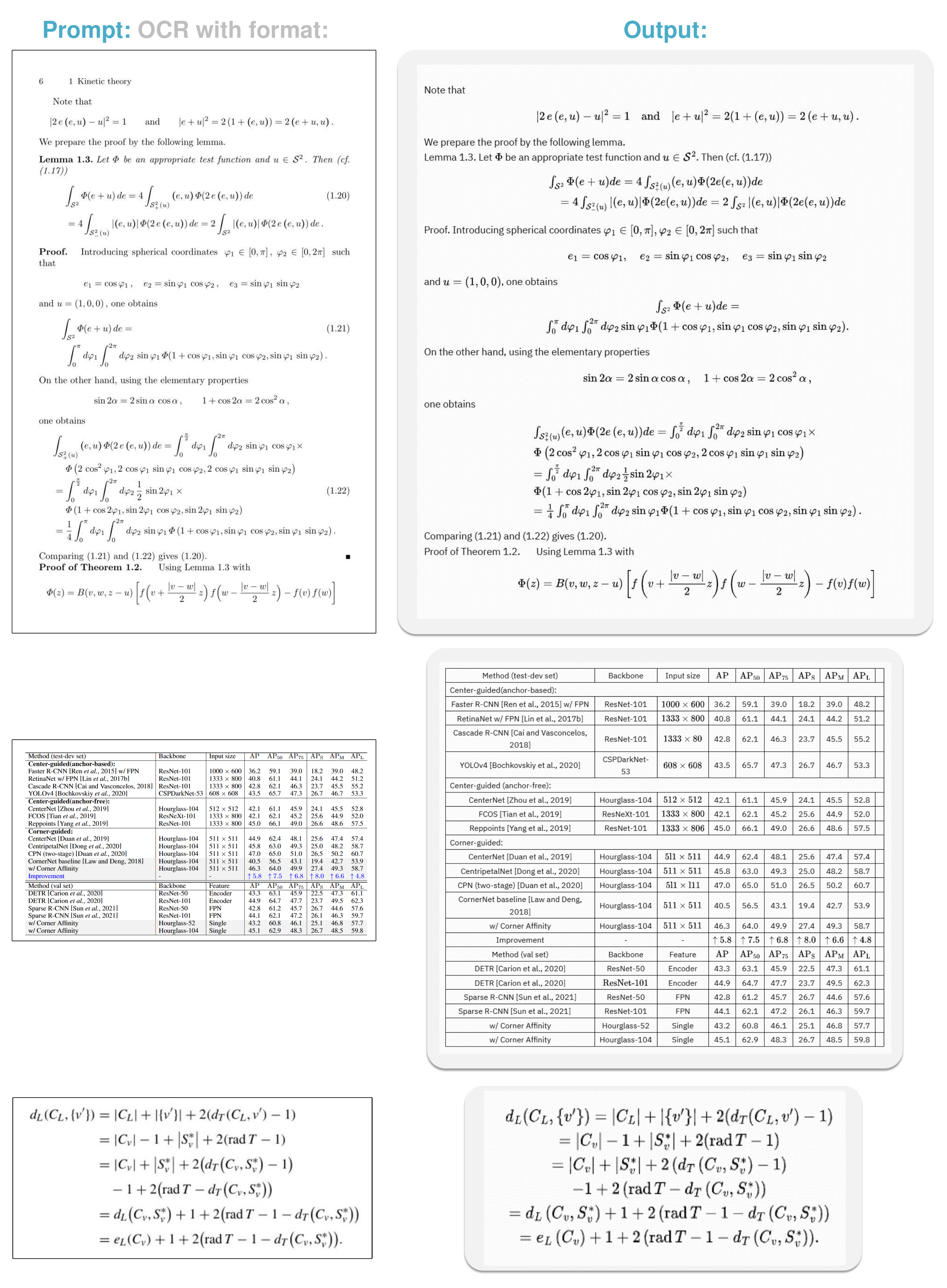}
\caption{The formatted text OCR ability of GOT. GOT works well on full-page texts and table/formula slice texts. These input forms are the most commonly used in document OCR, which proves that GOT has great prospects in application.}
\label{fig:app1}
\end{figure}

\begin{figure}[!t]
\centering
\includegraphics[width=1.0\textwidth]{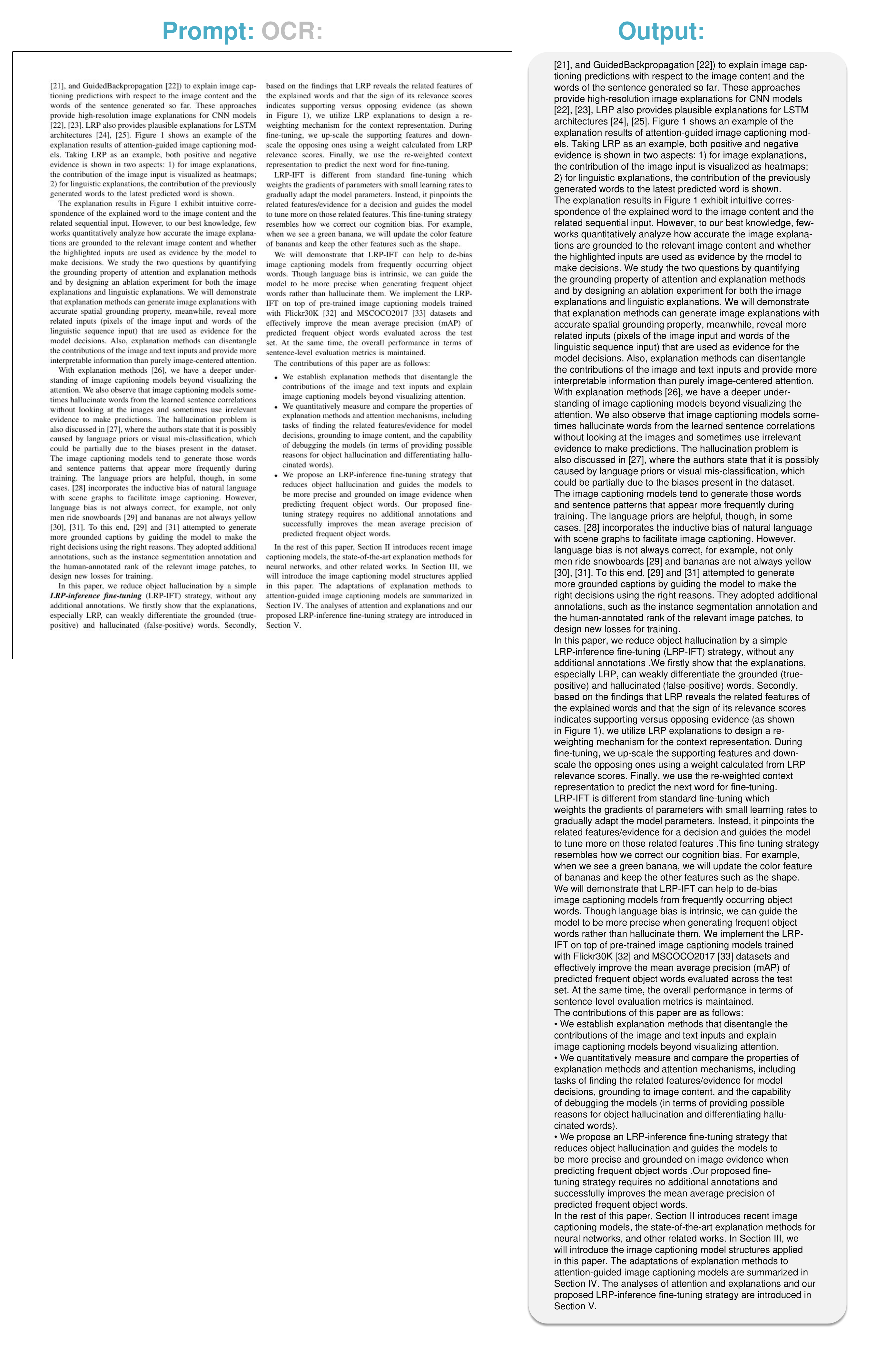}
\caption{The plain text (document) OCR ability of GOT. For double-column documents with high text density, GOT can still handle them well, proving the excellent text perception ability.}
\label{fig:app2}
\end{figure}

\begin{figure}[!t]
\centering
\includegraphics[width=1.0\textwidth]{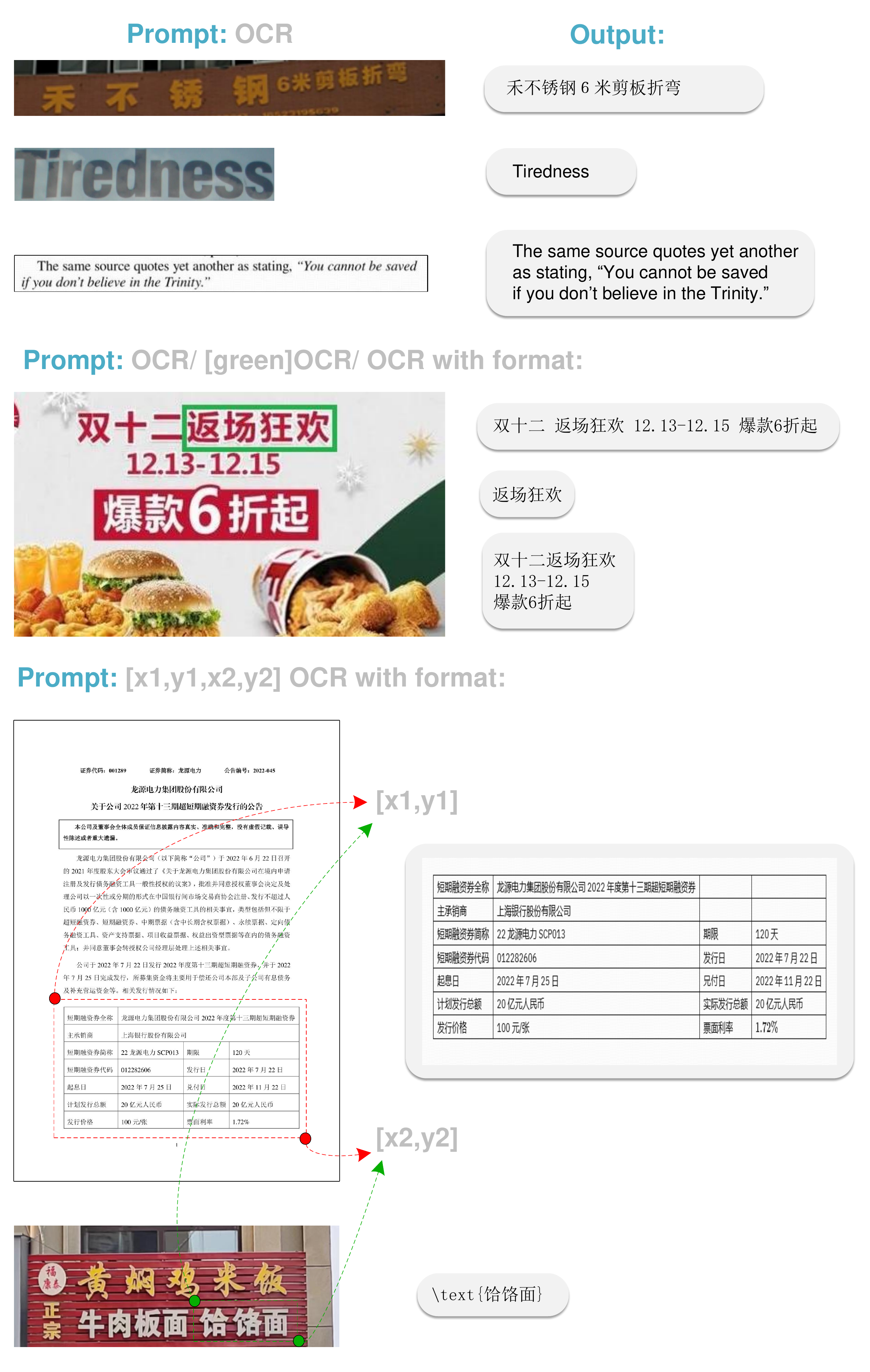}
\caption{Scene OCR and fine-grained OCR results of GOT.  We equip GOT with more interactive fine-grained OCR tasks, allowing it to output OCR results of regions of interest based on prompts.}
\label{fig:app3}
\end{figure}

\begin{figure}[!t]
\centering
\includegraphics[width=1.0\textwidth]{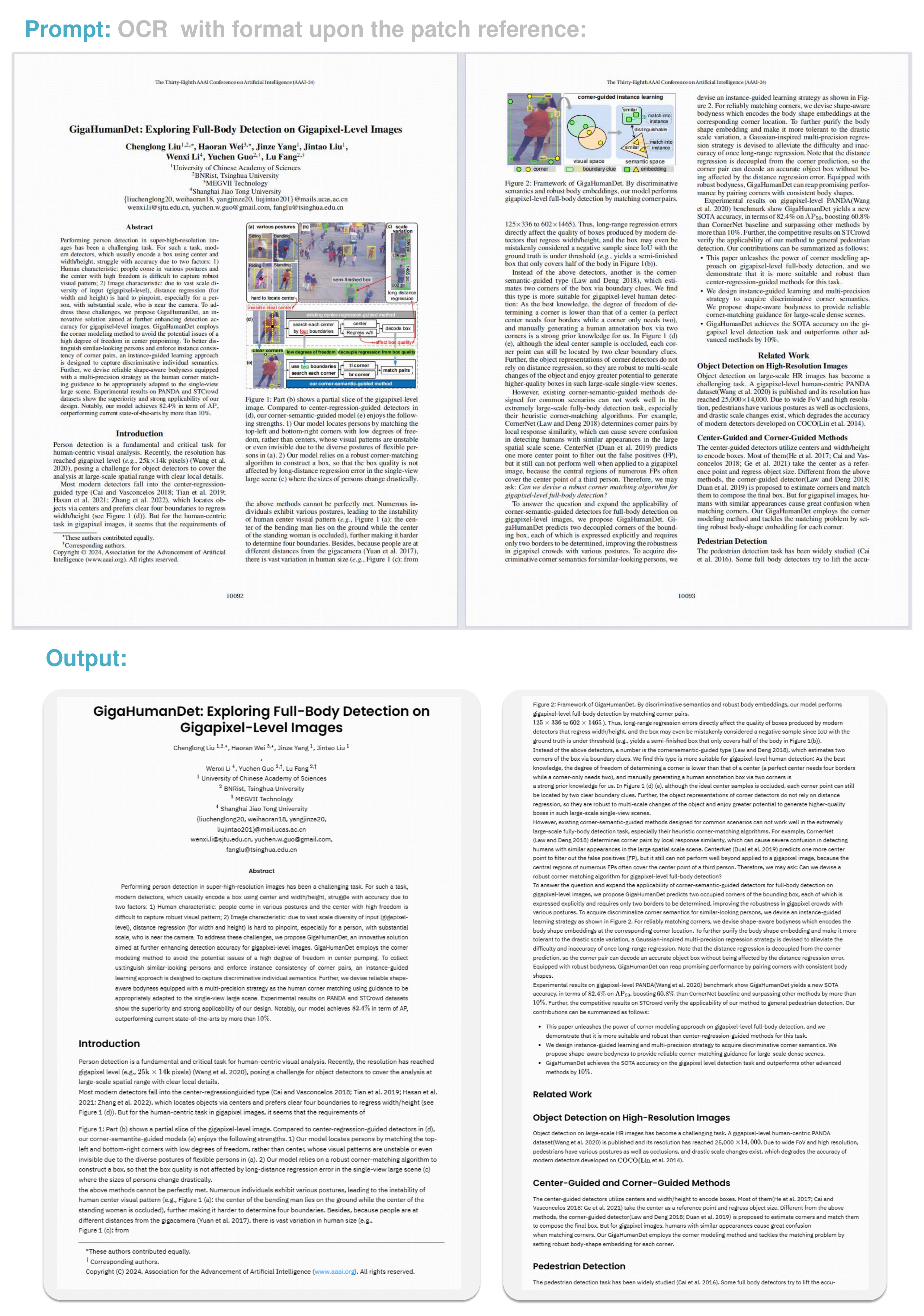}
\caption{Dynamic resolution of GOT for high-resolution images. In the dual-page paper reading mode shown in the figure (from \cite{liu2024gigahumandet}), the input resolution of the original GOT is not sufficient to handle it. Therefore, we adapt dynamic resolution technology to make the model no longer limited to the size of the image.}
\label{fig:app4}
\end{figure}

\begin{figure}[!t]
\centering
\includegraphics[width=0.85\textwidth]{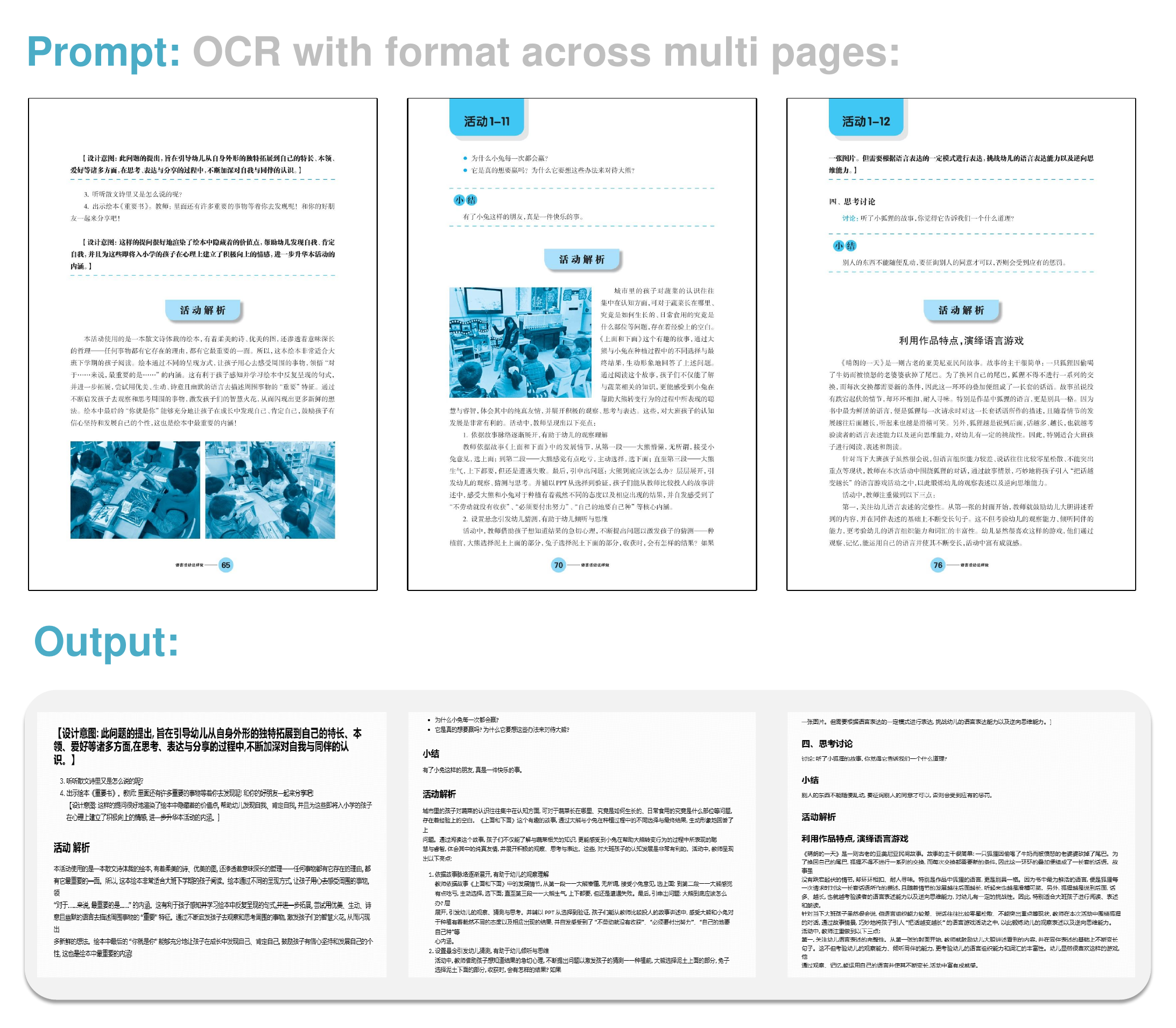}
\caption{Multi-page (document) OCR ability of GOT. With this feature, researchers can continue to train the GOT with multi-page PDF-text pairs, such as Arxiv paper with \textit{.tex} file.}
\label{fig:app5}
\end{figure}

\begin{figure}[!h]
\centering
\includegraphics[width=0.8\textwidth]{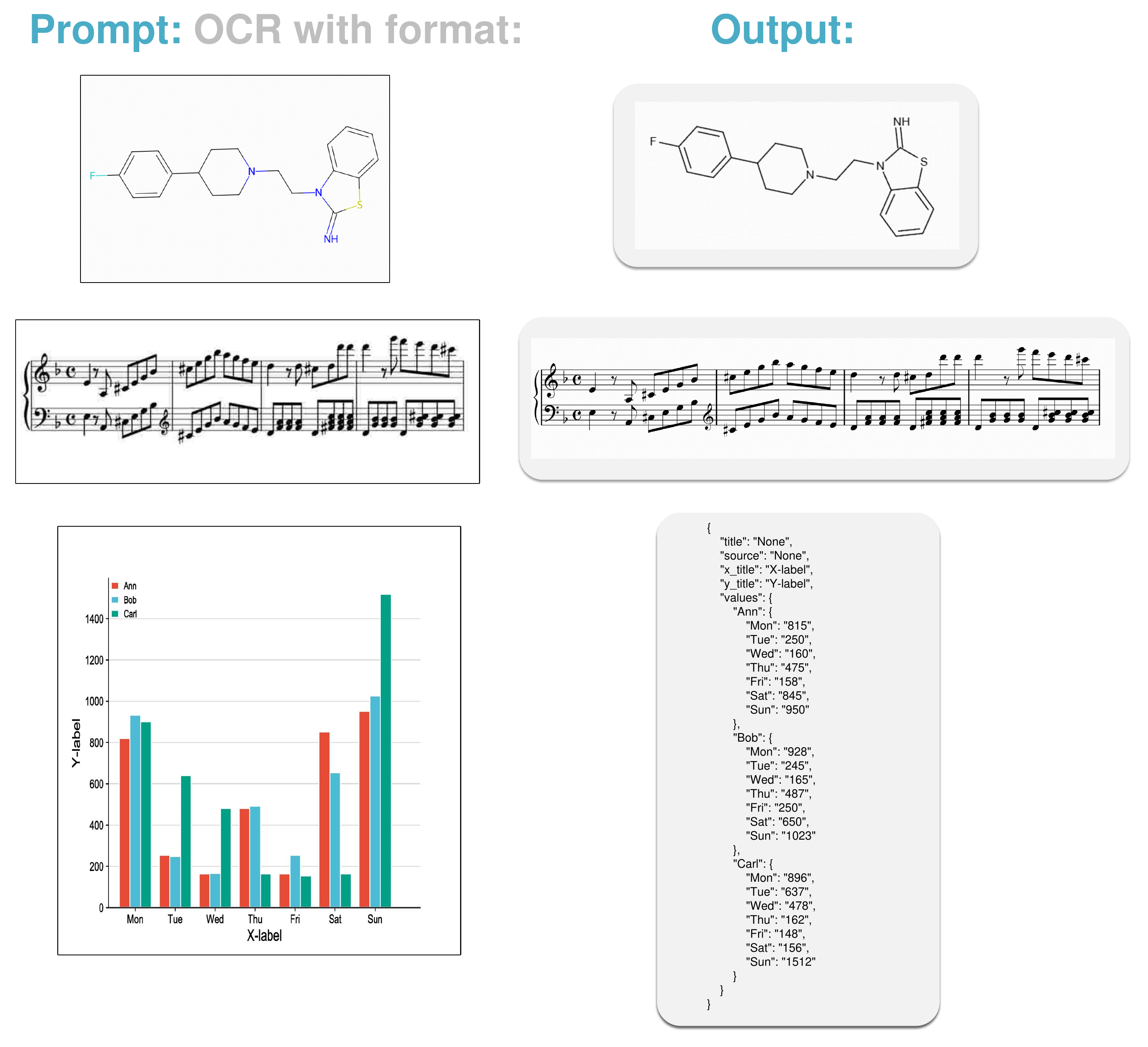}
\caption{More general OCR results. GOT can process molecular formulas, sheet music, and charts.  }
\label{fig:app6}
\end{figure}

\begin{figure}[!t]
\centering
\includegraphics[width=1\textwidth]{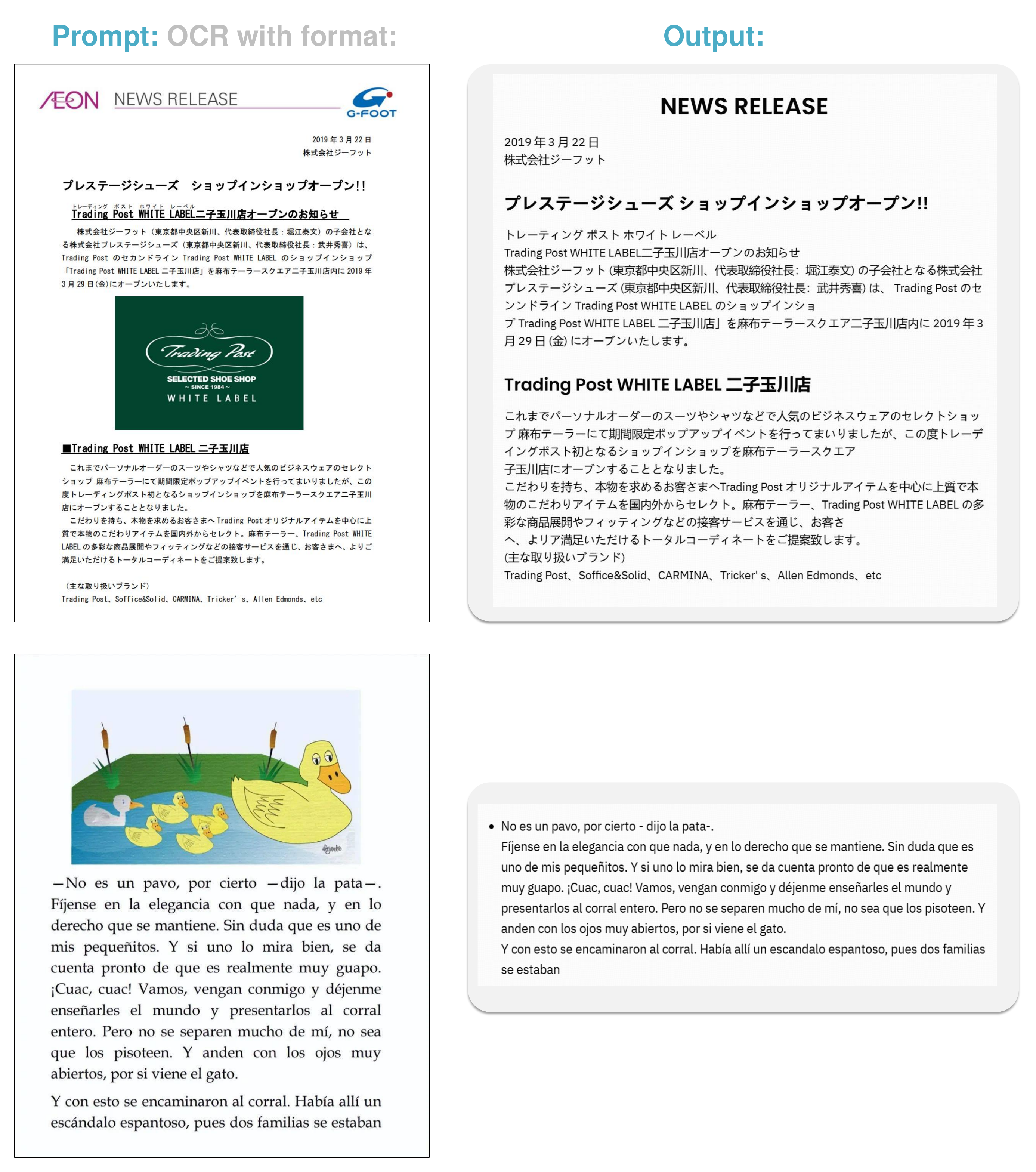}
\caption{We do not specifically introduce additional OCR capabilities for GOT other than Chinese and English. Yet the PDF data we crawled may contain a small amount of text in other languages, leading to the GOT seeming to have the ability to recognize other languages. However, we cannot guarantee the OCR quality of other languages. Therefore, we recommend fine-tuning the model with corresponding data if this feature is needed.}
\label{fig:app7}
\end{figure}

\clearpage

{
\small
\bibliographystyle{splncs04}
\bibliography{egbib}

\begin{thebibliography}{10}
\providecommand{\url}[1]{\texttt{#1}}
\providecommand{\urlprefix}{URL }
\providecommand{\doi}[1]{https://doi.org/#1}

\bibitem{Teklia_CASIA-HWDB2-line}
Casia-hwdb2-line. \url{https://huggingface.co/datasets/Teklia/CASIA-HWDB2-line} (2024)

\bibitem{Teklia_IAM-line}
Iam-line. \url{https://huggingface.co/datasets/Teklia/IAM-line} (2024)

\bibitem{Teklia_NorHand-v3-line}
Norhand-v3-line. \url{https://huggingface.co/datasets/Teklia/NorHand-v3-line} (2024)

\bibitem{qwen}
Bai, J., Bai, S., Chu, Y., Cui, Z., Dang, K., Deng, X., Fan, Y., Ge, W., Han, Y., Huang, F., Hui, B., Ji, L., Li, M., Lin, J., Lin, R., Liu, D., Liu, G., Lu, C., Lu, K., Ma, J., Men, R., Ren, X., Ren, X., Tan, C., Tan, S., Tu, J., Wang, P., Wang, S., Wang, W., Wu, S., Xu, B., Xu, J., Yang, A., Yang, H., Yang, J., Yang, S., Yao, Y., Yu, B., Yuan, H., Yuan, Z., Zhang, J., Zhang, X., Zhang, Y., Zhang, Z., Zhou, C., Zhou, J., Zhou, X., Zhu, T.: Qwen technical report. arXiv preprint arXiv:2309.16609  (2023)

\bibitem{Qwen-VL}
Bai, J., Bai, S., Yang, S., Wang, S., Tan, S., Wang, P., Lin, J., Zhou, C., Zhou, J.: Qwen-vl: A versatile vision-language model for understanding, localization, text reading, and beyond. arXiv preprint arXiv:2308.12966  (2023)

\bibitem{blecher2023nougat}
Blecher, L., Cucurull, G., Scialom, T., Stojnic, R.: Nougat: Neural optical understanding for academic documents. arXiv preprint arXiv:2308.13418  (2023)

\bibitem{calvo2020understanding_OMR}
Calvo-Zaragoza, J., Jr, J.H., Pacha, A.: Understanding optical music recognition. ACM Computing Surveys (CSUR)  \textbf{53}(4),  1--35 (2020)

\bibitem{chen2024onechart}
Chen, J., Kong, L., Wei, H., Liu, C., Ge, Z., Zhao, L., Sun, J., Han, C., Zhang, X.: Onechart: Purify the chart structural extraction via one auxiliary token. arXiv preprint arXiv:2404.09987  (2024)

\bibitem{chen2024far_intervl1.5}
Chen, Z., Wang, W., Tian, H., Ye, S., Gao, Z., Cui, E., Tong, W., Hu, K., Luo, J., Ma, Z., et~al.: How far are we to gpt-4v? closing the gap to commercial multimodal models with open-source suites. arXiv preprint arXiv:2404.16821  (2024)

\bibitem{paddleocrv2_du2021pp}
Du, Y., Li, C., Guo, R., Cui, C., Liu, W., Zhou, J., Lu, B., Yang, Y., Liu, Q., Hu, X., et~al.: Pp-ocrv2: Bag of tricks for ultra lightweight ocr system. arXiv preprint arXiv:2109.03144  (2021)

\bibitem{graves2006connectionist}
Graves, A., Fernández, S., Gomez, F., Schmidhuber, J.: Connectionist temporal classification: Labelling unsegmented sequence data with recurrent neural networks. In: International Conference on Machine Learning (ICML) (2006)

\bibitem{gu2022wukong}
Gu, J., Meng, X., Lu, G., Hou, L., Minzhe, N., Liang, X., Yao, L., Huang, R., Zhang, W., Jiang, X., et~al.: Wukong: A 100 million large-scale chinese cross-modal pre-training benchmark. Advances in Neural Information Processing Systems  \textbf{35},  26418--26431 (2022)

\bibitem{hu2024mplugdocowl1.5}
Hu, A., Xu, H., Ye, J., Yan, M., Zhang, L., Zhang, B., Li, C., Zhang, J., Jin, Q., Huang, F., et~al.: mplug-docowl 1.5: Unified structure learning for ocr-free document understanding. arXiv preprint arXiv:2403.12895  (2024)

\bibitem{lecun1998gradient_mnist}
LeCun, Y., Bottou, L., Bengio, Y., Haffner, P.: Gradient-based learning applied to document recognition. Proceedings of the IEEE  \textbf{86}(11),  2278--2324 (1998)

\bibitem{BLIP2}
Li, J., Li, D., Savarese, S., Hoi, S.: Blip-2: Bootstrapping language-image pre-training with frozen image encoders and large language models. arXiv preprint arXiv:2301.12597  (2023)

\bibitem{li2023trocr}
Li, M., Lv, T., Chen, J., Cui, L., Lu, Y., Florencio, D., Zhang, C., Li, Z., Wei, F.: Trocr: Transformer-based optical character recognition with pre-trained models. In: Proceedings of the AAAI Conference on Artificial Intelligence. vol.~37, pp. 13094--13102 (2023)

\bibitem{li2022exploring_vitdet}
Li, Y., Mao, H., Girshick, R., He, K.: Exploring plain vision transformer backbones for object detection. In: European conference on computer vision. pp. 280--296. Springer (2022)

\bibitem{liao2017textboxes}
Liao, M., Shi, B., Bai, X., Wang, C., Lu, T., Mei, T.: Textboxes: A fast text detector with a single deep neural network. In: Proceedings of the Thirty-First AAAI Conference on Artificial Intelligence (2017)

\bibitem{liao2022real_dbnet}
Liao, M., Zou, Z., Wan, Z., Yao, C., Bai, X.: Real-time scene text detection with differentiable binarization and adaptive scale fusion. IEEE transactions on pattern analysis and machine intelligence  \textbf{45}(1),  919--931 (2022)

\bibitem{liu2024focus_fox}
Liu, C., Wei, H., Chen, J., Kong, L., Ge, Z., Zhu, Z., Zhao, L., Sun, J., Han, C., Zhang, X.: Focus anywhere for fine-grained multi-page document understanding. arXiv preprint arXiv:2405.14295  (2024)

\bibitem{liu2024gigahumandet}
Liu, C., Wei, H., Yang, J., Liu, J., Li, W., Guo, Y., Fang, L.: Gigahumandet: Exploring full-body detection on gigapixel-level images. In: Proceedings of the AAAI Conference on Artificial Intelligence. vol.~38, pp. 10092--10100 (2024)

\bibitem{liu-2022-deplot}
Liu, F., Eisenschlos, J.M., Piccinno, F., Krichene, S., Pang, C., Lee, K., Joshi, M., Chen, W., Collier, N., Altun, Y.: Deplot: One-shot visual language reasoning by plot-to-table translation. In: Findings of the 61st Annual Meeting of the Association for Computational Linguistics (2023), \url{https://arxiv.org/abs/2212.10505}

\bibitem{liu2024llavanext}
Liu, H., Li, C., Li, Y., Li, B., Zhang, Y., Shen, S., Lee, Y.J.: Llava-next: Improved reasoning, ocr, and world knowledge (January 2024), \url{https://llava-vl.github.io/blog/2024-01-30-llava-next/}

\bibitem{llava}
Liu, H., Li, C., Wu, Q., Lee, Y.J.: Visual instruction tuning (2023)

\bibitem{liu2019icdar_ReCTS}
Liu, X., Zhang, R., Zhou, Y., Jiang, Q., Song, Q., Li, N., Zhou, K., Wang, L., Wang, D., Liao, M., et~al.: Icdar 2019 robust reading challenge on reading chinese text on signboard. arXiv preprint arXiv:1912.09641  (2019)

\bibitem{liu2019curved}
Liu, Y., Jin, L., Zhang, S., Luo, C., Zhang, S.: Curved scene text detection via transverse and longitudinal sequence connection. Pattern Recognition  \textbf{90},  337--345 (2019)

\bibitem{liu2024textmonkey}
Liu, Y., Yang, B., Liu, Q., Li, Z., Ma, Z., Zhang, S., Bai, X.: Textmonkey: An ocr-free large multimodal model for understanding document. arXiv preprint arXiv:2403.04473  (2024)

\bibitem{loshchilov2016sgdr}
Loshchilov, I., Hutter, F.: Sgdr: Stochastic gradient descent with warm restarts. arXiv preprint arXiv:1608.03983  (2016)

\bibitem{AdamW}
Loshchilov, I., Hutter, F.: Decoupled weight decay regularization. In: {ICLR} (2019)

\bibitem{lyu2018multi}
Lyu, P., Yao, C., Wu, W., Yan, S., Bai, X.: Multi-oriented scene text detection via corner localization and region segmentation. In: Proceedings of the IEEE conference on computer vision and pattern recognition. pp. 7553--7563 (2018)

\bibitem{masry2023unichart}
Masry, A., Kavehzadeh, P., Do, X.L., Hoque, E., Joty, S.: Unichart: A universal vision-language pretrained model for chart comprehension and reasoning. arXiv preprint arXiv:2305.14761  (2023)

\bibitem{masry2022chartqa}
Masry, A., Long, D.X., Tan, J.Q., Joty, S., Hoque, E.: Chartqa: A benchmark for question answering about charts with visual and logical reasoning. arXiv preprint arXiv:2203.10244  (2022)

\bibitem{DocVQA}
Mathew, M., Karatzas, D., Jawahar, C.: Docvqa: A dataset for vqa on document images. In: Proceedings of the IEEE/CVF winter conference on applications of computer vision. pp. 2200--2209 (2021)

\bibitem{mertz2007graphics_tikz}
Mertz, A., Slough, W.: Graphics with tikz. The PracTEX Journal  \textbf{1},  1--22 (2007)

\bibitem{methani2020plotqa}
Methani, N., Ganguly, P., Khapra, M.M., Kumar, P.: Plotqa: Reasoning over scientific plots. In: Proceedings of the IEEE/CVF Winter Conference on Applications of Computer Vision. pp. 1527--1536 (2020)

\bibitem{GPT4}
OpenAI: Gpt-4 technical report (2023)

\bibitem{radford2021learning}
Radford, A., Kim, J.W., Hallacy, C., Ramesh, A., Goh, G., Agarwal, S., Sastry, G., Askell, A., Mishkin, P., Clark, J., et~al.: Learning transferable visual models from natural language supervision. In: International conference on machine learning. pp. 8748--8763. PMLR (2021)

\bibitem{rios2024sheet_music}
R{\'\i}os-Vila, A., Calvo-Zaragoza, J., Paquet, T.: Sheet music transformer: End-to-end optical music recognition beyond monophonic transcription. arXiv preprint arXiv:2402.07596  (2024)

\bibitem{rios2023end_GrandStaff}
R{\'\i}os-Vila, A., Rizo, D., I{\~n}esta, J.M., Calvo-Zaragoza, J.: End-to-end optical music recognition for pianoform sheet music. International Journal on Document Analysis and Recognition (IJDAR)  \textbf{26}(3),  347--362 (2023)

\bibitem{schuhmann2022laion5b}
Schuhmann, C., Beaumont, R., Vencu, R., Gordon, C., Wightman, R., Cherti, M., Coombes, T., Katta, A., Mullis, C., Wortsman, M., et~al.: Laion-5b: An open large-scale dataset for training next generation image-text models. Advances in Neural Information Processing Systems  \textbf{35},  25278--25294 (2022)

\bibitem{shi2017icdar2017_RCTW}
Shi, B., Yao, C., Liao, M., Yang, M., Xu, P., Cui, L., Belongie, S., Lu, S., Bai, X.: Icdar2017 competition on reading chinese text in the wild (rctw-17). In: 2017 14th iapr international conference on document analysis and recognition (ICDAR). vol.~1, pp. 1429--1434. IEEE (2017)

\bibitem{TextVQA}
Singh, A., Natarajan, V., Shah, M., Jiang, Y., Chen, X., Batra, D., Parikh, D., Rohrbach, M.: Towards vqa models that can read. In: Proceedings of the IEEE/CVF conference on computer vision and pattern recognition. pp. 8317--8326 (2019)

\bibitem{tian2016detecting}
Tian, Z., Huang, W., He, T., He, P., Qiao, Y.: Detecting text in natural image with connectionist text proposal network. In: European conference on computer vision. pp. 56--72. Springer (2016)

\bibitem{veit2016coco_text}
Veit, A., Matera, T., Neumann, L., Matas, J., Belongie, S.: Coco-text: Dataset and benchmark for text detection and recognition in natural images. arXiv preprint arXiv:1601.07140  (2016)

\bibitem{wang2020contournet}
Wang, Y., Xie, H., Zha, Z.J., Xing, M., Fu, Z., Zhang, Y.: Contournet: Taking a further step toward accurate arbitrary-shaped scene text detection. In: Proceedings of the IEEE/CVF Conference on Computer Vision and Pattern Recognition. pp. 11753--11762 (2020)

\bibitem{wei2023vary}
Wei, H., Kong, L., Chen, J., Zhao, L., Ge, Z., Yang, J., Sun, J., Han, C., Zhang, X.: Vary: Scaling up the vision vocabulary for large vision-language models. arXiv preprint arXiv:2312.06109  (2023)

\bibitem{wei2024small_varytoy}
Wei, H., Kong, L., Chen, J., Zhao, L., Ge, Z., Yu, E., Sun, J., Han, C., Zhang, X.: Small language model meets with reinforced vision vocabulary. arXiv preprint arXiv:2401.12503  (2024)

\bibitem{xia2024chartx}
Xia, R., Zhang, B., Ye, H., Yan, X., Liu, Q., Zhou, H., Chen, Z., Dou, M., Shi, B., Yan, J., Qiao, Y.: Chartx \& chartvlm: A versatile benchmark and foundation model for complicated chart reasoning (2024)

\bibitem{ye2023mplug}
Ye, J., Hu, A., Xu, H., Ye, Q., Yan, M., Dan, Y., Zhao, C., Xu, G., Li, C., Tian, J., et~al.: mplug-docowl: Modularized multimodal large language model for document understanding. arXiv preprint arXiv:2307.02499  (2023)

\bibitem{ye2023ureader}
Ye, J., Hu, A., Xu, H., Ye, Q., Yan, M., Xu, G., Li, C., Tian, J., Qian, Q., Zhang, J., et~al.: Ureader: Universal ocr-free visually-situated language understanding with multimodal large language model. arXiv preprint arXiv:2310.05126  (2023)

\bibitem{zhang2019shopsign}
Zhang, C., Peng, G., Tao, Y., Fu, F., Jiang, W., Almpanidis, G., Chen, K.: Shopsign: A diverse scene text dataset of chinese shop signs in street views. arXiv preprint arXiv:1903.10412  (2019)

\bibitem{zhang2021adaptive}
Zhang, S.X., Zhu, X., Yang, C., Wang, H., Yin, X.C.: Adaptive boundary proposal network for arbitrary shape text detection. In: Proceedings of the IEEE/CVF International Conference on Computer Vision. pp. 1305--1314 (2021)

\bibitem{OPT}
Zhang, S., Roller, S., Goyal, N., Artetxe, M., Chen, M., Chen, S., Dewan, C., Diab, M., Li, X., Lin, X.V., et~al.: Opt: Open pre-trained transformer language models. arXiv preprint arXiv:2205.01068  (2022)

\bibitem{zhong2019publaynet}
Zhong, X., Tang, J., Yepes, A.J.: Publaynet: largest dataset ever for document layout analysis. In: 2019 International conference on document analysis and recognition (ICDAR). pp. 1015--1022. IEEE (2019)

\bibitem{zhou2017east}
Zhou, X., Yao, C., Wen, H., Wang, Y., Zhou, S., He, W., Liang, J.: East: An efficient and accurate scene text detector. In: Proceedings of the IEEE International Conference on Computer Vision (ICCV) (2017)

\end{thebibliography}
}


\end{document}